\documentclass[conference]{IEEEtran}
\IEEEoverridecommandlockouts
\usepackage{cite}
\usepackage{amsthm}
\usepackage{amsmath,amssymb,amsfonts}
\usepackage{algorithmic}
\usepackage{graphicx}
\usepackage{textcomp}
\usepackage{xcolor}
\usepackage{makecell}
\usepackage{booktabs}

\theoremstyle{definition}
\newtheorem{definition}{Definition}

\def\BibTeX{{\rm B\kern-.05em{\sc i\kern-.025em b}\kern-.08em
    T\kern-.1667em\lower.7ex\hbox{E}\kern-.125emX}}
\begin{document}

\setlength{\abovedisplayskip}{2pt}
\setlength{\belowdisplayskip}{2pt}

\title{DynGraph2Seq: Dynamic-Graph-to-Sequence Interpretable Learning for Health Stage Prediction in Online Health Forums}

\author{\IEEEauthorblockN{Yuyang Gao}
\IEEEauthorblockA{\textit{George Mason University} \\
ygao13@gmu.edu}
\and
\IEEEauthorblockN{Lingfei Wu}
\IEEEauthorblockA{\textit{IBM Research} \\
wuli@us.ibm.com}
\and
\IEEEauthorblockN{Houman Homayoun}
\IEEEauthorblockA{\textit{George Mason University} \\
hhomayou@gmu.edu}
\and
\IEEEauthorblockN{Liang Zhao}
\IEEEauthorblockA{\textit{George Mason University} \\
lzhao9@gmu.edu}

}

\maketitle

\begin{abstract}
Online health communities such as the online breast cancer forum enable patients (i.e., users) to interact and help each other within various subforums, which are subsections of the main forum devoted to specific health topics. The changing nature of the users' activities in different subforums can be strong indicators of their health status changes. This additional information could allow health-care organizations to respond promptly and provide additional help for the patient. However, modeling complex transitions of an individual user's activities among different subforums over time and learning how these correspond to his/her health stage are extremely challenging. In this paper, we first formulate the transition of user activities as a dynamic graph with multi-attributed nodes, then formalize the health stage inference task as a dynamic graph-to-sequence learning problem, and hence propose a novel dynamic graph-to-sequence neural networks architecture (DynGraph2Seq) to address all the challenges. Our proposed DynGraph2Seq model consists of a novel dynamic graph encoder and an interpretable sequence decoder that learn the mapping between a sequence of time-evolving user activity graphs and a sequence of target health stages. We go on to propose dynamic graph hierarchical attention mechanisms to facilitate the necessary multi-level interpretability. A comprehensive experimental analysis of its use for a health stage prediction task demonstrates both the effectiveness and the interpretability of the proposed models.
\end{abstract}


\section{Introduction}
% online forum - anonymous (advantage)
Online healthcare forums and communities \cite{BCC, ACS, ehealth} such as the Breast Cancer Community have greatly changed the way patients seek health-related information and have become an important part of patients' lives.
% Unlike traditional approaches, where patients only receive information about their disease from their care providers, these online forums and communities have enabled millions of patients to ask questions related to their diseases, interact with other patients with similar prognoses, and provide support to each other across the world. 
The communications and interactions between patients in online forums can provide valuable information about a patient's emotional well-being and behaviors related to the management of their health that conventional clinical data collected from hospital information systems and electronic health records (EHR) is unable to capture. 
% Thus, the online healthcare forum data act as a great complement resource to the clinical data on getting a full understanding of patients health condition.
% Moreover, beyond conventional online communities and social media, online health communities provide a unique way to analyze and infer patients' health stages and disease history. 
% Figure \ref{fig:signature} shows an example of a patient's health stage information extracted from a patient signature that contains the cancer diagnosis and treatment history, along with the relevant dates.
The synergies between the information on patients' online communication and health status make possible a unique and wide range of research topics on health informatics \cite{elhadad2014characterizing, zhang2014does ,zhang2017longitudinal} that rely on both patients' interactions in online forums as well as their health stage records. 

\begin{figure}
\centering
\includegraphics[width=0.8\linewidth]{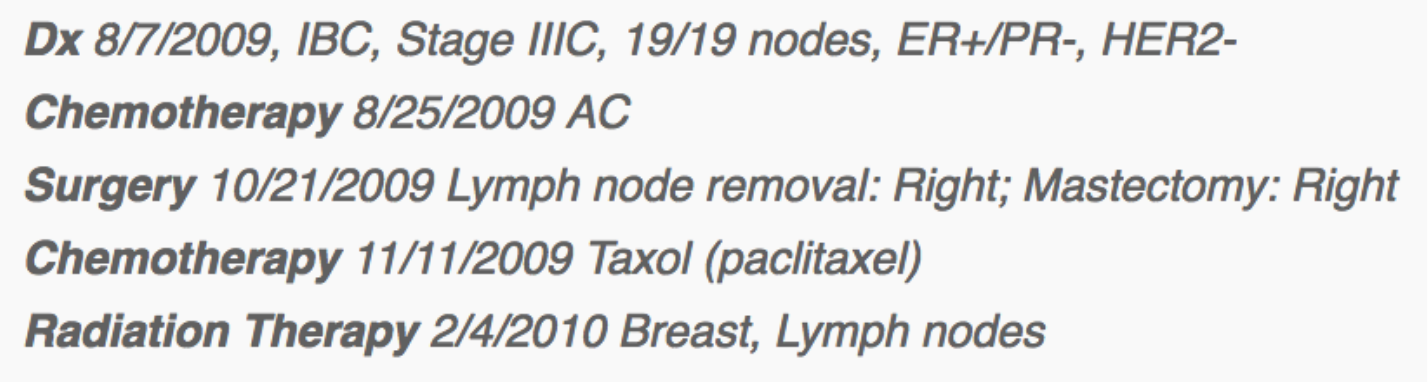}
\vspace{-10pt}
\caption{An example of patient signature that contains cancer diagnosis and treatment history.}
\label{fig:signature}
\vspace{-20pt}
\end{figure}

\begin{figure}
\centering
\vspace{-0.3cm}
\includegraphics[width=\linewidth]{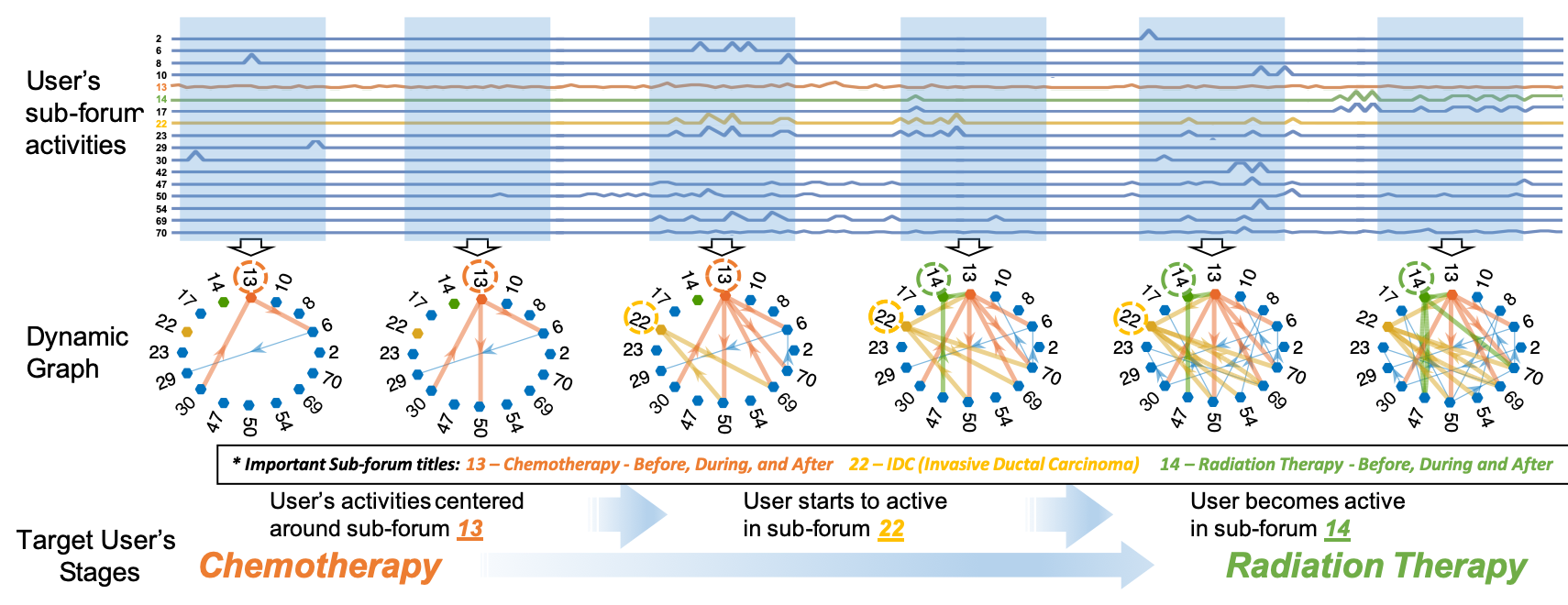}\vspace{-0.3cm}
\caption{An example of user forum activities and the corresponding health stage evolution. 
In the first two time windows, the user is mainly active in Subforum $\#13$ while going through chemotherapy treatment. In the third time window, the user starts to be active in Subforum $\#22$ at about the time when she undergoes IDC treatment. Finally, in the last three time windows, the user becomes active in Subforum $\#14$ when she enters the ``Radiation Therapy'' health treatment stage.
}
\label{fig:formulation}\vspace{-0.5cm}
\end{figure}

However, the health stage information in the online health community has some unique challenges and characteristics.
First, though some patients share their disease history, as shown in Figure \ref{fig:signature}, such information is not provided or is simply missing for many others. 
For instance, over 36\% active users that registered within recent 2 years have not yet shared their disease history in the Breast Cancer Community.
% Important information about patients' health stages is highly desirable to be able to infer or predict these patients' health stage information. 
Second, different subforums under specific topics are often correlated to specific disease stages. 
For example, in the online breast cancer forum, the patients who are active in the ``Chemotherapy - Before, During, and After'' subforum typically look for information related to their Chemotherapy treatment. 
% Thus, users' activities within these subforums could serve as a strong indicator of an individual user's current health stage. 
Third, as the patients' health conditions progress over time, they often move from one set of subforums to others that are more related to their new health stages. 
Therefore, for each patient, these transitions among subforums can lead to an inter-connected subforum activity network that evolves over time, which could be highly entangled with the progress of patient's health status, as shown in Figure \ref{fig:formulation}.

The ability to accurately infer users' missing health stage information is crucial, as this could enable health care organizations to better support patients by pinpointing the most valuable information for each at their particular health stage \cite{jha2010cancer}.
To infer the missing user health stage information, the correspondence between the users' forum activities and their health stage history needs to be accurately identified and modeled. 
Naturally, the networked and time-evolving forum activity data can be formulated as a dynamic sequence of user activity transition graphs that change over time.
In addition, the target user health stage history can be formulated as a sequence that needs to be inferred. 
Thus, without loss of generality, a new generic task is presented here where the goal is to learn the mapping from a sequence of graph-structured data to a target sequence. 
In this paper, we limit our scope to the domain of online health forums and focus on health stage sequence prediction based on online health forums data.

However, capturing the high-level mapping between the evolution of the user activity networks and the changes in the corresponding user's health stage can be very difficult due to the following challenges:
\textbf{1) Difficulty in modeling the forum data, which is dynamic, networked, and multi-attributed.} A user's activities in the various subforums can change dynamically over time and these activity transitions naturally bridge different subforums. 
\textbf{2) Difficulty in learning the association between a sequence of user activity networks and the corresponding sequence of health stages}. The sequence of user activity networks contains complicated graph-structured information that dynamically evolves over time. Developing end-to-end learning between such dynamic complex data and a specific sequence is highly difficult.
\textbf{3) Lack of interpretability of the health stage sequence inference process.} 
The sequence of user activity networks has a two-level hierarchical structure, namely node (i.e., subforum) to network level, and network to health stage level. 
It is thus a major objective to incorporate this hierarchical structural information into the development of an interpretable health stage inference process. 

In this paper, we formally define the generic learning problem of health stage sequence inference using online forum data and propose the first framework to address the aforementioned challenges effectively. The contribution of this paper is four-fold: 1) we define the health stage inference problem in online health forums and formulate the user activities as transition graphs that are capable of modeling user dynamic transitions between subforums and their complex relationships; 2) we propose a novel deep neural encoder-decoder framework for learning the mapping between complex dynamic graph sequence inputs and the target output sequence; 3) we propose a new dynamic graph hierarchical attention mechanism that captures both the time-level and node-level attention, thus providing model transparency throughout the whole inference process; 4) experiments on online health forum dataset demonstrate that our proposed models outperform conventional sequence inference methods. In addition, our qualitative analyses and case studies provide interpretable insights into the learning results of the proposed model and its variations.
\section{Related work}
Our model draws inspiration from the research fields of online health community analysis, dynamic graph learning, attention mechanisms, and neural encoder-decoder models.
\subsection{Online Health Communities Analysis}
A number of studies have focused on the analysis and utilization of online health communities data. Popular social media is good for aggregate level pattern mining tasks \cite{zhao2016hierarchical, wang2018multi}. However, their power is limited for discovering individual-level health stages and health network patterns due to the privacy issues involved and data scarcity. There have been several analyses of breast cancer forum data \cite{elhadad2014characterizing, zhang2014does} and, more recently, machine learning models have been used for longitudinal analysis \cite{zhang2017longitudinal} and some binary classification tasks\cite{jha2010cancer}. However, we are the first to propose a general framework that can achieve health stage sequence inference using online forum data.

\subsection{Dynamic Graph Representation Learning}
As an emerging topic in the graph representation learning domain, dynamic graph learning has attracted a great deal of attention from researchers in recent years \cite{trivedi2018representation, zhou2018dynamic, goyal2018dyngraph2vec}. 
However, these graph embedding techniques typically focus on learning representations of the graphs, such as node embedding, but in many real-world applications the aim is to learn some high-level knowledge from the graph data, such as graph classification tasks \cite{wu2018multiple,Wu:2019:SGA:3292500.3330918} and graph to sequence tasks \cite{xu2018sql, xu2018exploiting}. An end-to-end learning model is thus needed to learn the mapping between the whole sequence of graph data and the target output sequence, instead of merely focusing on learning node representations.

\subsection{Attention Mechanism}
The attention mechanism was first proposed by \cite{bahdanau2014neural} and has been widely used for machine translation tasks \cite{luong2015effective, yang2016hierarchical}. 
The attention mechanism has also been introduced in the graph representation learning domain \cite{velickovic2017graph, yang2018graph}. However, there is little to no work that focuses specifically on studying the unique hierarchical structure that is naturally present in dynamic graphs.

\subsection{Neural Encoder-Decoder Models}
The neural encoder-decoder models \cite{cho2014learning,bahdanau2014neural} have been widely extended to model the mapping of general object inputs to their corresponding sequences \cite{eriguchi2016tree,tai2015improved}.
Recent advances in graph deep learning and graph convolutional networks have enabled various graph deep learning models to handle challenges in the domains of graph generation \cite{guo2018deep, simonovsky2018graphvae, li2018learning} and graph-to-sequence learning \cite{xu2018graph2seq}. However, there have been no reports of work that explores dynamic graph to sequence learning, where the natural sequential order contained in a dynamic graph and its sequences might be advantageous for neural encoder-decoder models.

\section{Problem Formulation}

%online Forum data->dynamic graph

\subsection{User Forum Activities as a Dynamic Graph}

% The online forum data records the path of each user's transition from one subforum to another, as well as their activities within each subforum. In order to capture these complex transitions and model the relationships between subforums, we propose a novel method to formulate the raw user subforum activities into activity transition networks that preserve these characteristics.

An activity transition network is formulated naturally as follows. User activities are first partitioned into a series of time windows. We then begin by formulating a node for each subforum, with a transition from one forum to the other deemed to occur if the most active forum (based on visiting time or number of postings) switches from the former to the latter, creating a directed `edge' between them. Each node (i.e., subforum) also records the user activity in the forum to build the activity transition network. 
% For example in Figure \ref{fig:formulation}, the subforum transition sequence is $\{30 \rightarrow 13 \rightarrow 6 \rightarrow 29\}$, where 30, 13, 6, and 29 are the IDs of the subforums visited. Thus, the transition edges for the first snapshot graph will be (30,13), (13,6), and (6,29). The graph in each time window records all the transitions in and previous to it.
Naturally, such time-ordered activity transition networks can be formally defined as dynamic graphs, also known as temporal networks in the network science literature \cite{li2017fundamental}, that capture the complex dynamic characteristics and time-evolving features of graphs, as defined in the following.

\theoremstyle{definition}
\begin{definition}{(dynamic graph).}
\textit{A dynamic graph $\mathcal{G}=\{G_1, G_2,$ $\cdots, G_T\}$ is an ordered sequence of $t=1,\cdots, T$ separate graphs on the same set of $|V|=N$ nodes, with each snapshot graph $G_t(V,E_t)$ characterized by a weighted adjacency matrix $A_t\in\mathbb{R}^{N \times N}$ and a set of node features $F_t \in\mathbb{R}^{N \times D}$ for a given time window, where $D$ represents the total number of node features.}
\end{definition}

% node features in online forum data

We can now formulate the activity transition networks as a dynamic graph, illustrated in Figure \ref{fig:formulation}. Here, the dynamic graph contains a sequence of snapshot graphs $G_1, G_2, \cdots, G_6$ that characterize user activities in the online forum for a given time period, where $G_t$ represents the snapshot graph $G_t(V,E_t)$ for simplicity. Each node $v\in V$ represents a subforum devoted to a specific topic and the edges $E_t$ capture the user's movement between different subforums at a given time window shown as blue boxes. Each node $v$ contains a set of features $F_{t,v}$ that represents the topics covered by the specific subforum. By formulating user activities as dynamic graphs, the mapping between the evolution of the user activity and the changes of user's health stages will be preserved. 

% machine learning model now will be capable of capturing the dynamic changes of user's subforum activities from subforum 13 (node and edges are marked as orange color), which is a subforum mainly talks about ``Chemotherapy'' topics, to subforum 22 (node and edges are marked as yellow color), which contains topics about ``Invasive Ductal Carcinoma (IDC)'', until the user reaches a new health treatment stage ``Radiation Therapy'' and the activities finally centered at  ``Radiation'' related sub-form 14 (node and edges are marked as green color). Thus,
\subsection{Learning Sequence from Dynamic Graph}
As we can see from Figure \ref{fig:formulation}, there is a clear mapping between the evolution of the user activity dynamic graph and changes in the corresponding user's health stage. 
Motivated by this observation, we can formulate such problems as a general dynamic graph to sequence problem as follows:

Given a dynamic graph $\mathcal{G}=\{G_1, G_2, \cdots, G_T\}$ as input data, the goal is to predict the target sequence $S=\{s_1, s_2, \cdots, s_M\}$, where $s_m \in \mathbb{V}$ is the $m$th token of the output sequence in vocabulary $\mathbb{V}$; and $T$ and $M$ are the input graph sequence length and output sequence length, respectively. 
Formally, this problem is equivalent to learning a translation mapping from input dynamic graph $\mathcal{G}$ to a sequence $S$ as $\{G_1, G_2, \cdots, G_T\} \rightarrow \{s_1, s_2, \cdots, s_M\}$.

The translation mapping problem between some source objects and target sequences has been widely studied, including both graph-to-sequence \cite{xu2018graph2seq} and sequence-to-sequence \cite{sutskever2014sequence, cho2014learning} formulations. However, dynamic-graph-to-sequence translation is more complex and poses several unique challenges, namely 1) Difficulty in comprehensively modeling the dynamic multi-attributed network-structured data, as both complex relationships and dynamic evolving characteristics need to be captured; 2) The temporal dependency of snapshot graphs in the dynamic graph need to be modeled and constrained by the learning model; and 3) The learned translation mapping is often obscure and hard to explain or verify. This is because the original low-level representation (i.e. the node level at a specific time) is aggregated into the high-level representation (i.e. the dynamic graph as a whole), making it much more difficult to backtrack and explain the correspondence.
%Those challenges prevent current techniques from handling the task.

\begin{figure} 
\centering
\vspace{-0.7cm}
\includegraphics[width=\linewidth]{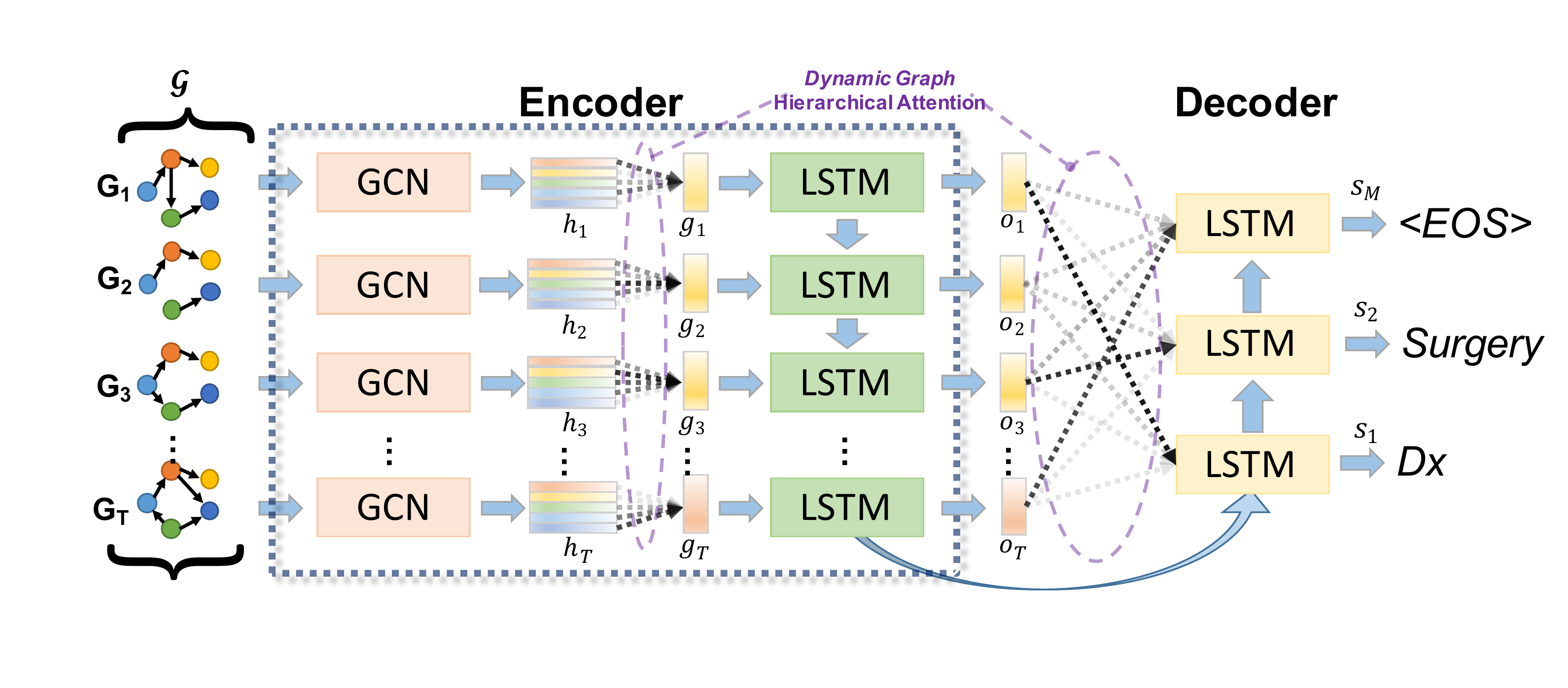}\vspace{-0.9cm}
\caption{
The proposed end-to-end dynamic graph-to-sequence learning (DynGraph2Seq) framework.
It includes a novel dynamic graph encoder and a sequence decoder with dynamic graph hierarchical attention. 
% The proposed framework not only generates sequence outputs by capturing complicated interactions of user's activities and dynamic characteristics of the evolving graphs, but also provides time-level and subforum level interpretability of the correlations between user's activities and health stages through the proposed two-level attention mechanism.
}
\label{fig:model}\vspace{-0.5cm}
\end{figure}

\section{Dynamic Graph-To-Sequence Model}

% \subsection{The DynGraph2Seq framework}
% In this section, we introduce our dynamic graph-to-sequence framework that includes a novel dynamic graph encoder and a sequence decoder, as shown in Figure \ref{fig:model}. To the best of our knowledge, this is the first end-to-end dynamic graph-to-sequence learning framework. Our DynGraph2Seq framework not only generates sequence outputs by capturing complicated interactions between a user's activities and the dynamic characteristics of the evolving graphs over time, but also provides both time-level and subforum level interpretability of the correlations between a user's online forum activities and that user's current health stages through our two-level attention mechanisms.

\subsection{Dynamic Graph Encoder}
The base model of our graph convolutional network for each snapshot graph is inspired by graph2seq \cite{xu2018graph2seq}, which was originally proposed for addressing static graph-to-sequence learning problems. The Graph2Seq model employs an inductive node embedding algorithm that generates bi-directional node embeddings by aggregating information from a node local forward and backward neighborhood within $k$ hops for a static graph. We extend this idea for dynamic graphs by applying such graph convolution on each snapshot graph within dynamic graph inputs. Specifically, suppose the total number of hops is $k$, then the hidden representation of $n$-th node in the snapshot graph $G_t$ after applying the first graph convolutional layer will be computed as follows:
\begin{gather}
\label{eq:node_embedding_1}
h^{\vdash}_{t,n}=mean(\{\sigma(W^{\vdash(1)}_{t}F_{t,u} + b^{\vdash(1)}_{t}), u\in \mathcal {N_{\vdash}}(v)\}) \\
h^{\dashv}_{t,n}=mean(\{\sigma(W^{\dashv(1)}_{t}F_{t,u} + b^{\dashv(1)}_{t}), u\in \mathcal{N_{\dashv}}(v)\}) \\
h_{t,n}^{(1)}=concat[h^{\vdash}_{t,n}, h^{\dashv}_{t,n}]
\end{gather}
where $\mathcal {N_{\vdash}} (v)$ represents the set of forward neighbor nodes of node $v$, whereas $\mathcal {N_{\dashv}} (v)$ represents the set of backward neighbor nodes; $W^{\dashv(1)}_{t}, b^{\dashv(1)}_{t}$ and $W^{\vdash(1)}_{t}, b^{\vdash(1)}_{t}$ are learnable parameters for the first convolution layer. $F_{t,u}$ is the feature vector of node $u$ in a snapshot graph at time step $t$; $\sigma(\cdot)$ represents the activation function of the network (e.g. ReLU); the $mean(\cdot)$ function takes the element-wise mean of the set of vectors in the equation; and $concat[vec1,vec2]$ concatenates the two row vectors into a single row vector.

Likewise, for hop $k$, the hidden representation of the $n$-th node in the snapshot graph $G_t$ can be computed via the hidden representations computed from layer $k-1$.
% , as follows:
% \begin{gather}
% \label{eq:node_embedding_2}
% h^{\vdash}_{t,n}=mean(\{\sigma(W^{\vdash(k)}_{t}h_{t,u}^{(k-1)} + b^{\vdash(k)}_{t}), u\in \mathcal {N_{\vdash}}(v)\}) \\
% h^{\dashv}_{t,n}=mean(\{\sigma(W^{\dashv(k)}_{t}h_{t,u}^{(k-1)} + b^{\dashv(k)}_{t}), u\in \mathcal{N_{\dashv}}(v)\}) \\
% h_{t,n}^{(k)}=concat[h^{\vdash}_{t,n}, h^{\dashv}_{t,n}]
% \end{gather}
Finally, after applying $k$ layers of convolutions, the final hidden representation of the $n$-th node in the snapshot graph $G_t$ will be output as $h_{t,n}=h_{t,n}^{(k)}$. 
% \begin{gather}
% h_{t,n}=h_{t,n}^{(k)}
% \end{gather}

In order to capture the high-level representation of graphs for end-to-end graph learning, aggregating node level embeddings to graph level embedding that conveys the entire graph information is essential. To achieve this, we adopt the max pooling operation proposed by \cite{xu2018graph2seq, wu2018multiple} as the base aggregation function, which feeds the node embeddings $h_{t,n}$ to a fully-connected layer and then applies the max pooling method element-wise for each snapshot graph $G_t$ to yield a sequence of graph-level representations $g_t$.
To model the graph dynamic changes and long-term dependencies throughout the $M$ steps, we utilize Long Short Term Memory (LSTM) networks \cite{hochreiter1997long} as a graph embedding sequence encoder to learn the entire dynamic graph-level embedding. 
% The computation of the LSTM network at time step $t$ is defined as:
% \begin{gather}
% f_{t}=\sigma(W_f \cdot [o_{t-1},g_t] + b_f) \\
% in_t=\sigma(W_{in} \cdot [o_{t-1},g_t] + b_{in}) \\
% \widetilde C_t=tanh(W_C \cdot [o_{t-1},g_t] + b_C) \\
% C_t=f * C_{t-1} + in_t * \widetilde C_t \\
% out_t=\sigma(W_{out} \cdot [o_{t-1},g_t] + b_{out}) \\
% o_t=out_t * tanh(C_t)
% \end{gather}
% where $o_t$ is the output of the LSTM network at time step $t$, $C_t$ is the new cell state for the next time step computation, and the initial cell state for the encoder is set to all-zeros.

\subsection{Sequence Decoder with Dynamic Graph Hierarchical Attention}

Once the dynamic graph encoder takes the sequence of snapshot graphs $G_t$ and aggregates node embeddings to generate a sequence of graph-level embeddings that capture the entire dynamic graph's global characteristics, the LSTM layer will output the final hidden-state of encoder $C_T$ to summarize all the graph-level embeddings. Then, in the sequence decoding phase, we utilize a conventional sequence decoder \cite{luong17} and set the initial cell state of the decoder as $C_T$ in order to decode the target sequence $S$.

However, there are two issues with this simple sequence decoder: 1) the effectiveness of the sequence decoder depends on the length of the dynamic graph sequence; and 2) the predicted user's health stage sequence need to be interpretable based on the dynamic graph sequence at both the time-level and node-level . 
% For instance, as shown in Figure \ref{fig:formulation}, our model must learn to pinpoint which snapshot graphs in the dynamic graph sequence are strongly correlated to the output user health stage predicted by the decoder. To take this one step deeper, the model should also be able to provide information on which of the important nodes (subforums) are in a given snapshot graph while the model is generating a specific user's health stage.
To handle the above questions pertaining to model interpretability,
% we need to develop a more effective way of handling information propagation and aggregation from low-level representations (i.e. node levels at a specific time) to high-level representations (i.e. the dynamic graph as a whole). 
we propose a novel dynamic graph hierarchical attention mechanism that includes \textbf{node-to-graph} and \textbf{graph-to-sequence} attention that is capable of enhancing the interpretability for node embedding aggregation and capture the hierarchical structure of user online forum activities over time more effectively. 

\begin{figure}
\vspace{-0.2cm}
\centering
\includegraphics[width=0.5\linewidth]{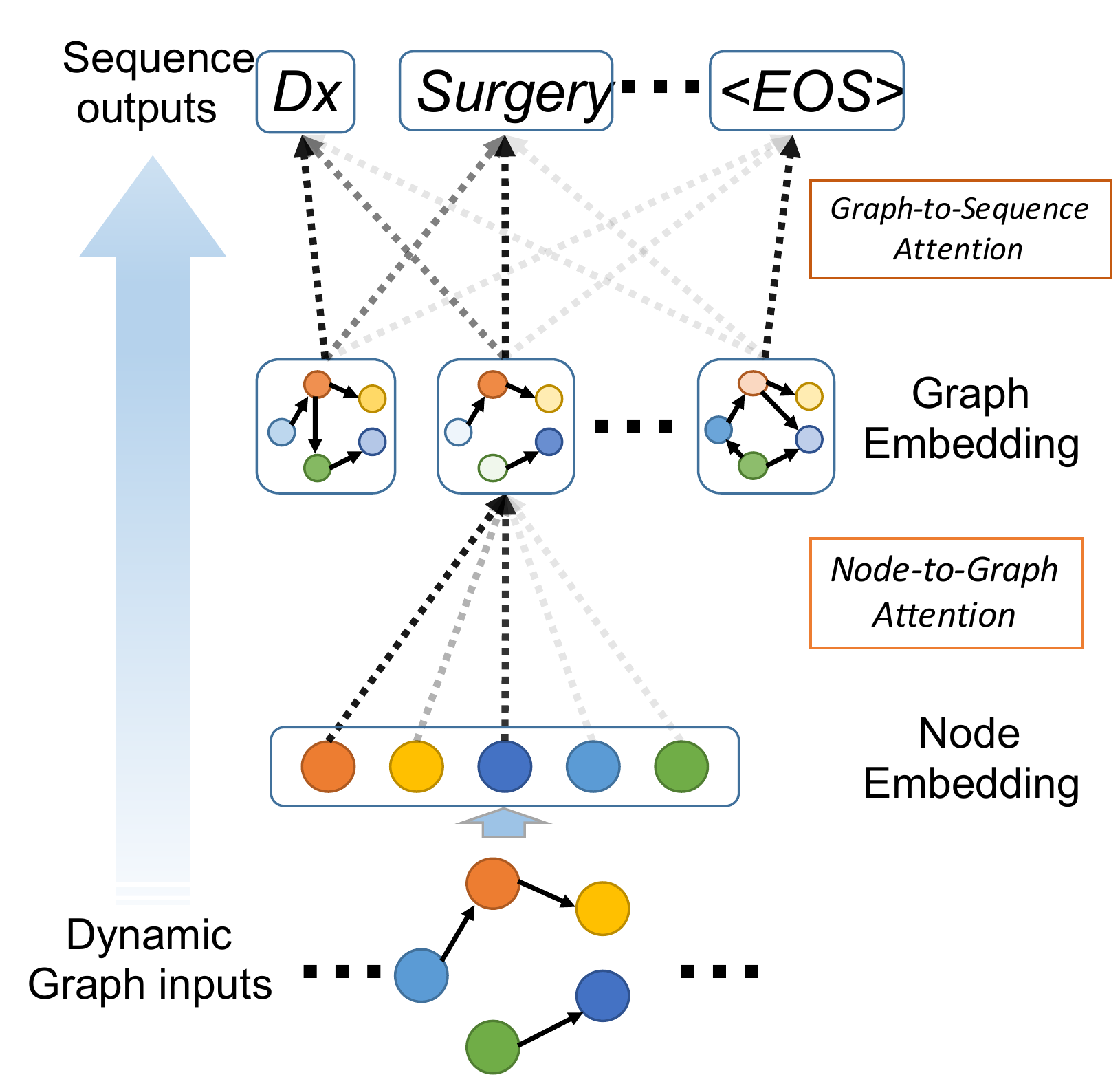}\vspace{-0.3cm}
\caption{The proposed dynamic graph hierarchical attention mechanism: node-to-graph and graph-to-sequence attention
% The node-to-graph attention aggregates the node level information (i.e. node embeddings) to formulate the graph-level embeddings, and the graph-to-sequence attention aims to find the mapping between each snapshot graph and each token in output sequence.
}\vspace{-0.6cm}
\label{fig:HAT}
\end{figure}

\subsubsection{Node-to-Graph Attention}
\hfill\\
Once the node embeddings of a graph have been computed, an average or max pooling operation \cite{xu2018graph2seq, wu2018multiple} is typically employed as the base aggregation function to obtain the graph-level embedding for the current graph. Although this works well in their individual settings, it does not work properly in our case since not all node embeddings contribute equally to the representation of the graph. For example, although a patient may view multiple subforums within a given time period, only a few important subforums will be correlated with the specific health stage of the patient. Therefore, it is vital to identify these important nodes (subforums) that contribute most to representing the embedding of the current graph. 
Inspired by \cite{raffel2015feed}, we adopt the feed-forward attention to aggregate the node embeddings and formulate the graph-level embeddings. Figure \ref{fig:HAT} shows an example of how the node-to-graph attention is computed for a snapshot graph $G_t$. For a given snapshot graph at step $t$, the \emph{node-to-graph attention} is given as follows:
\begin{gather}
\label{eq:node_to_graph_attention}
e_{t,n}\!\!=a(h_{t,n}) \quad
\!\!\alpha_{t,n}\!\!=\frac{\exp{(e_{t,n})}}{\sum\nolimits_{k=1}^{N} \exp{(e_{t,k})}} \quad
\!\!g_t=\sum\nolimits_{n=1}^{N} \!\!\alpha_{t,n} h_{t,n} \notag
\end{gather}
where the function $a(\cdot)$ is a learnable function that depends on the node embeddings $h_{t,n}$; and $g_t$ denotes the aggregated graph-level embedding for a snapshot graph at step $t$. In this formulation, the attention weights $\alpha_{t,n}$ explicitly model the importance of each node $n$ when constructing the graph-level representation of $g_t$. Clearly, we can utilize the attention weight information for each node to pinpoint which nodes (subforums) are highly related to the current health stage. We will discuss the interpretability of our node-to-graph attention in detail in the experimental Section.

\subsubsection{Graph-to-Sequence Attention}
\hfill\\
Once the graph-level embedding $g_t$ has been obtained for each snapshot graph $G_t$, the whole sequence of graph embeddings $g = \{g_i\}_{i=1}^T$ is fed into the sequence decoder, which generates the global hidden embedding $c$ that characterizes the entire sequence of dynamic graph information. Following the conventional encoder-decoder setup, $c$ is set as the initial hidden state for the sequence decoder from which to generate the target sequence of the health stages.

Although the hidden vector $c$ theoretically contains all the information needed for generating the target sequence, the encoder's hidden representation $o_t$ also contains valuable information about the snapshot graph information at that time step during the sequence encoding.
To reward such snapshot graphs, we use the attention mechanism and introduce graph-to-sequence level attention to measure the importance of each snapshot graph with the target sequence. Specifically, as shown in Figure \ref{fig:HAT}, the graph-to-sequence attention takes the sequence of hidden states for each graph $o=\{o_1,\cdots,o_T\}$ in the dynamic graph sequence as additional inputs to the decoder. This forces the decoder to consider both the current hidden state and the attention alignments between each word generated and for the whole sequence $o$. 

% Therefore, suppose the decoder is at time step $i$ and the hidden state of the previous step is represented as $s_{i-1}$, our graph-to-sequence attention is computed as follows:
% \vspace{-0.1cm}
% \begin{gather}
% \label{eq:mono_graph_to_graph_seq_attention}
% e_{i,t}=a(s_{i-1},o_t) \quad p_{i,t}=\sigma (e_{i,t})\\
% q_{i,t}=(1-p_{i,t-1})\cdot q_{i,t-1} + \alpha_{i-1,t} \\
% \alpha_{i,t}=p_{i,t}q_{i,t} \quad r_i=\sum\nolimits_{t=1}^{T} \alpha_{i,t} o_t
% \end{gather}
% where $q_{i,0}=0$ and $p_{i,0}=0$ for computing for the special case when $t=1$; the context vector $r_i$ is then used to compute the current hidden state in the decoder and generate a word in the target sequence.

\section{Experiments}
We evaluated the performance of our proposed model utilizing a real-world online health forum, namely the \textit{breast cancer community}. All the experiments were conducted on a 64-bit machine with Intel(R) Xeon(R) W-2155 CPU \@3.30GHz processor, 32GB memory and an NVIDIA TITAN Xp GPU.

\begin{table*}
  \caption{Performance Evaluation for Health Stage Prediction. The scores were obtained from 20 individual runs and presented in a mean $\pm$ standard deviation (SD) format. }
  \vspace{-0.3cm}
  \label{tab:bleu}
  \centering
  \begin{tabular}{c|ccccc}
    \toprule
    Model & BLEU-1 & BLEU-2 & BLEU-3 & BLEU-4 & ROUGE \\
    \midrule
    NMT(seq2seq) (w/o att)          & 55.5$\pm$2.38 & 38.4$\pm$0.91 & 27.1$\pm$0.90 & 19.2$\pm$0.87 & 71.6$\pm$1.04\\
    NMT(seq2seq) (w/ att)           & 57.8$\pm$1.86 & 40.4$\pm$1.21 & 29.0$\pm$1.28 & 20.1$\pm$1.06 & 72.9$\pm$0.86\\
    \midrule
    Graph2Seq (w/o att)             & 57.5$\pm$1.72 & 41.5$\pm$0.94 & 29.8$\pm$0.72 & 20.3$\pm$0.85 & 75.8$\pm$1.20\\
    %[20.8, 20.8, 22.9]
    Graph2Seq (w/ att)              & 58.2$\pm$2.19 & 41.1$\pm$1.38 & 30.1$\pm$0.83 & 21.0$\pm$0.51 & 76.2$\pm$0.96\\
    %[23.2, 22.6, 22.9] 
    \midrule
    DynGraph2Seq (w/o att)   & 60.9$\pm$1.53 & 43.7$\pm$1.00 & 31.5$\pm$0.63 & 22.1$\pm$0.48 & 79.3$\pm$0.80\\
    % DynGraph2Seq (w/ reg)           & 61.5$\pm$2.42 & 45.1$\pm$1.86 & 32.3$\pm$1.31 & 23.1$\pm$1.05 & 78.5$\pm$0.86\\
    DynGraph2Seq (w/ att)           & 62.3$\pm$1.46 & 44.7$\pm$1.29 & 32.0$\pm$0.94 & 22.5$\pm$1.13 & 80.8$\pm$0.36\\
    % DynGraph2Seq (w/ reg \& att)    & 64.1$\pm$0.84 & 45.4$\pm$0.31 & 33.1$\pm$0.41 & 24.1$\pm$0.70 & 81.0$\pm$0.69\\
    \bottomrule
  \end{tabular}
  \vspace{-0.5cm}
\end{table*}

\subsection{Experimental Settings}
\textbf{Online Breast Cancer Community Dataset}: The Breast Cancer Community \cite{BCC} is one of the largest online forums designed for patients to share information related to breast cancer. The forum data collected for this study covers an 8 year period from the beginning of 2010 to the end of 2017. To create user subforum activity transition graph sequences, we defined user activities as being when they posted new topics or replied to existing topics and the time window was set as one month. After removing common words and stop words, we extracted the 100 top frequency keywords from the forum content to construct the feature vectors for the subforums. We randomly selected 70\% of users who provided their health stage history for training, another 10\% for validation, and the remaining 20\% for testing.
The predicted health stage sequences were validated against the real health stage history extracted from the users' signatures. The vocabulary of the health stages consists of $\{$`Dx'\footnote{Short for Oncotype DX test, an initial diagnosis that analyzes how a cancer is likely to behave and respond to treatment.}, `Chemotherapy', `Targeted', `Hormonal', `Radiation', `Surgery'$\}$. 
%An example of a user health stage history is: ``\textit{Dx Chemotherapy Surgery Radiation Hormonal Surgery}''. 

\subsubsection{Evaluation Metrics}
\hfill\\
We used BLEU scores \cite{papineni2002bleu} and ROUGE-1 score \cite{lin2004rouge} as evaluation metrics for determining the closeness of the model predicted health stage history and the ground truth.

\subsubsection{Comparison Methods}
\hfill\\
\textbf{NMT(seq2seq)} The Neural Machine Translation model \cite{luong17} is a widely used state-of-the-art sequence-to-sequence model for machine transition tasks. Since the NMT model can only handle simple sequence inputs, we simplified the input data by concatenating the transition sequences of user activity for each month together in time order. The subforum features are omitted in such formulations. 
\hfill\\
\textbf{Graph2seq} The Graph2seq model \cite{xu2018graph2seq} is a general-purpose encoder-decoder model for static graph to sequence learning. Since the model cannot handle dynamic graphs, we simplified the input by aggregating all the edges that appeared in the dynamic graph together into a single static graph.

\subsubsection{Hyper-parameter Settings}
\hfill\\
We used the Adam optimizer\cite{kingma2014adam} with a learning rate of 0.001 and a batch size of 50 for model training; greedy search was used for sequence decoders.
Hyper-parameters were searched based on the highest scores achieved on the validation set.

\subsection{Performance}
Table \ref{tab:bleu} shows the model performance of the baseline and proposed models. 
The scores were obtained from 20 individual runs and presented in a mean $\pm$ standard deviation (SD) format. 
In general, our proposed DynGraph2Seq framework significantly outperformed both the Seq2Seq and Graph2Seq baselines for the various model settings and evaluation metrics. 
The basic DynGraph2Seq framework with the proposed dynamic graph hierarchical attention achieved the best score on all the metrics, outperforming the baseline models by 7\% - 17\% on the BLEU scores and 6\% - 13\% on the ROUGE scores. The baseline Graph2Seq model also achieved good scores, but was not as competitive as our proposed model. This was largely because Graph2Seq model failed to capture the dynamic characteristics of user activity with only static graph inputs. The Seq2Seq model performed badly due to its inability to model the complex relationships between the subforums with simple sequence inputs.

% Interestingly, although the full version of DynGraph2Seq (i.e. with the proposed hierarchical attentions) largely outperformed the baselines, the base model only achieved a marginal improvement compared to the Graph2Seq model. This was likely due to the fact that the aforementioned challenges prevented the base model from being fully effective for the learning task. These results further demonstrate that the proposed dynamic graph hierarchical attention is essential if the framework is to handle the learning tasks effectively.

\subsection{Interpretablity Analysis}
\begin{figure}
\centering
\includegraphics[width=0.5\linewidth]{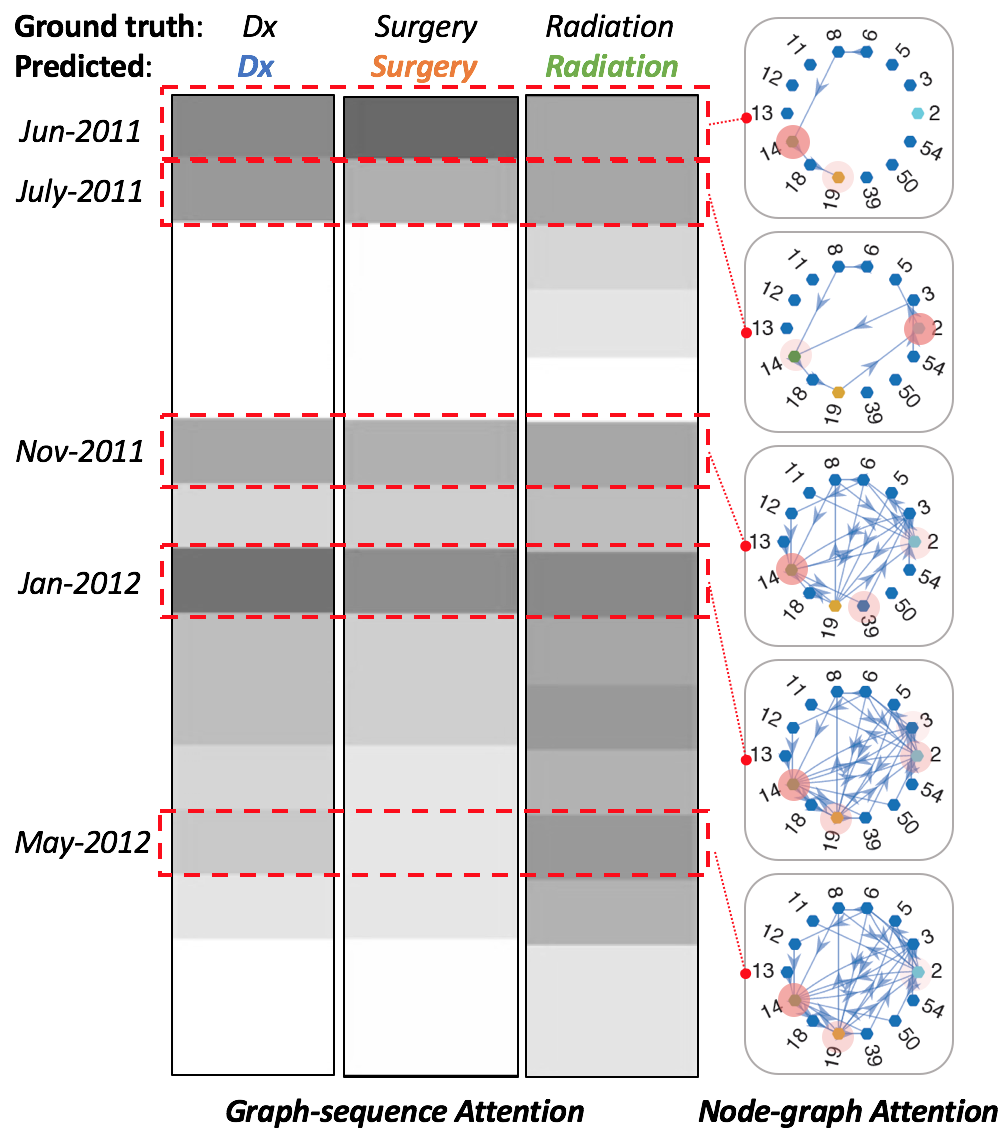}
\vspace{-0.4cm}
\caption{An example of learned dynamic graph hierarchical attention by DynGraph2Seq. The darker the color, the greater the attention being paid.}
\vspace{-0.5cm}
\label{fig:cs_HAT}
\end{figure}

Figure \ref{fig:cs_HAT} shows an example of the learned dynamic graph hierarchical attention by DynGraph2Seq. The left part of the figure shows the graph-to-sequence attention learned by the model, where each column is a grayscale heatmap representing the amount of attention being paid to each snapshot graph when the model predicted a specific health stage. The darker the color, the greater the attention being paid. We can see much attention was paid to the graphs around the months being labeled in the figure. The graphs for each labeled months are shown on the right. Interestingly, the graphs in the first two months attracted more attention from the model because those were the months when the patient first became active in the breast cancer online forum. The last two labeled snapshot graphs relate approximately to the time when the user engaged in extensive activities in a wide variety of subforums.

To understand why these particular snapshot graphs were important , we went one step deeper by examining the node-to-graph level attention. The red spots on the nodes shown on the right side of Figure \ref{fig:cs_HAT} represent the amount of attention being paid to each node (i.e. subforum). Again the darker the red spot, the greater the attention being paid. Now the attention becomes even more interesting and interpretable. For example, when constructing the representation of the May-2012 snapshot graph,  Subforum $\#14$ was assigned the most attention. The title of it is actually ``Radiation Therapy - Before, During and After'', which is strongly correlated to the health stage `Radiation'.
% This explains why that particular graph received more graph-to-sequence attention when the model predicted `Radiation'.
Likewise, we further discovered that Subforum $\#2$, entitled ``Not Diagnosed but Worried'', has a strong correlation with `Dx' and Subforum $\#19$, entitled ``DCIS (Ductal Carcinoma In Situ)'', is a strong indicator for `Surgery'. These observed correspondences confirm that the proposed dynamic graph hierarchical attention mechanism greatly enhances the interpretability of the model.

\section{Conclusion}
In this paper, we formulated the task of health stage inference using online health forum data as a dynamic graph-to-sequence learning problem and propose a novel DynGraph2Seq architecture that can handle this new type of learning problem effectively. 
Our DynGraph2Seq model consists of a novel dynamic graph encoder and an interpretable sequence decoder to learn the mapping between a sequence of time-evolving user activity graphs and a sequence of target health stages.
In addition, we developed a dynamic graph hierarchical attention to facilitate the multi-level interpretability.
Our comprehensive experiments and analyses for health stage prediction demonstrate both the effectiveness and the interpretability of the proposed models.

\section*{Acknowledgement}
This work was supported by the National Science Foundation grant: \#1755850, \#1841520, \#1907805, Jeffress Trust Award, and NVIDIA GPU Grant.

%
% The next two lines define the bibliography style to be used, and the bibliography file.
\bibliographystyle{./IEEEtran}
\bibliography{./sample-base}

\end{document}